\documentclass[journal]{IEEEtran}
\usepackage{makecell}
\usepackage{supertabular}
\usepackage{multirow}
\usepackage{graphicx}
\usepackage{subfigure}
\usepackage[normalem]{ulem}
\usepackage{url}
\usepackage{bbm}
\usepackage{enumerate}
\usepackage{longtable}
\usepackage{booktabs}
\usepackage{multirow}
\usepackage{xspace}

\usepackage{multicol}
\usepackage{color}
\usepackage{threeparttable}
\usepackage{ulem}
\usepackage{comment}

\usepackage{float}
\usepackage{listings}
\usepackage{caption}
\usepackage{subcaption}
\usepackage{algorithm,algpseudocode}
\usepackage{amsmath}
\usepackage{amsfonts}
\usepackage{amssymb}

\newcommand{\Name}{STEMS\xspace}

\usepackage{xcolor,soul,framed} 

\colorlet{shadecolor}{yellow}

\usepackage{array}
\usepackage{mdwmath}
\usepackage{mdwtab}
\usepackage{eqparbox}
\usepackage{url}

\begin{document}

    \title{ STEMS: Spatial-Temporal Enhanced Safe Multi-Agent Coordination for Building Energy Management}
    \author{Huiliang Zhang\thanks{Huiliang Zhang, Di Wu, and Benoit Boulet are with the Department of Electrical Computer Engineering, McGill University, Montreal, QC H3A 0G4, Canada (e-mail: huiliang.zhang2@mail.mcgill.ca; di.wu5@mcgill.ca; benoit.boulet@mcgill.ca).}, Di Wu,~\IEEEmembership{Member,~IEEE,} Arnaud Zinflou\thanks{Arnaud Zinflou is with the Data Science Team, Institut de Recherche Hydro Québec, Montreal, QC, Canada (e-mail: zinflou.arnaud@hydroquebec.com).},~\IEEEmembership{Senior Member,~IEEE,} and Benoit Boulet,~\IEEEmembership{Senior Member,~IEEE}}


\markboth{IEEE Internet of Things Journal
}{Zhang \MakeLowercase{\textit{et al.}}: }

\maketitle

\begin{abstract}
  Building energy management is essential for achieving carbon reduction goals, improving occupant comfort and reducing energy costs. 
  Coordinated building energy management faces critical challenges in exploiting spatial-temporal dependencies while ensuring operational safety across multi-building systems. Current multi-building energy systems face three key challenges: insufficient spatial-temporal information exploitation, lack of rigorous safety guarantees, and system complexity. 
  This paper proposes Spatial-Temporal Enhanced Safe Multi-Agent Coordination (STEMS), 
  a novel safety-constrained multi-agent reinforcement learning framework for coordinated building energy management.
  STEMS integrates two core components: (1) a spatial-temporal 
  graph representation learning framework using GCN-Transformer fusion architecture to capture inter-building relationships and temporal patterns, 
  and (2) a safety-constrained multi-agent RL algorithm incorporating Control Barrier Functions to provide mathematical safety guarantees. 
Extensive experiments on real-world building datasets demonstrate STEMS's strong performance over existing methods, showing that STEMS achieves 21\% cost reduction and 18\% emission reduction over traditional methods, 65\% safety violation reduction compared with learning-based baselines, and maintains the lowest discomfort rate.
  The framework also demonstrates strong robustness during extreme weather conditions and maintains effectiveness across different building types. 
\end{abstract}


\begin{IEEEkeywords}
Building energy management, multi-agent, reinforcement learning, spatial-temporal graph networks, safety constraints.
\end{IEEEkeywords}

%
\IEEEpeerreviewmaketitle


\section{Introduction}
Buildings consume approximately 40\% of global energy and account for 36\% of CO$_2$ emissions worldwide~\cite{chen2024green}.
This massive energy consumption directly impacts environmental sustainability and economic costs as urbanization accelerates.
Moreover, as urban residents spend over 80\% of their time indoors, 
building energy management directly affects daily life quality.
People require comfortable indoor environments under safe building control while facing rising energy costs and environmental concerns.
This creates a complex multi-objective optimization challenge where comfort, safety, economic efficiency, and environmental sustainability must be balanced simultaneously~\cite{zhang2022building-survey}.
Current building energy management systems operate independently among individual buildings, 
which limits their ability to achieve optimal performance~\cite{kumari2024multi}.

The widespread deployment of smart sensors and IoT devices has enabled buildings to monitor real-time conditions 
and share information about energy demand, storage status, and operational strategies.
This information sharing capability opens new ways for coordinated energy management across building clusters.
Reinforcement learning (RL) provides a promising solution for building energy management 
because it can learn optimal control policies without requiring precise system models.
Unlike traditional model predictive control (MPC) methods that rely on accurate mathematical models, 
RL algorithms can adaptively learn through interaction with the environment, 
handling system uncertainties and complexities effectively.
In multi-building scenarios, each building can act as an intelligent agent, 
enabling coordinated optimization through multi-agent reinforcement learning (multi-agent RL).
Multi-agent RL allows buildings to learn collaborative strategies while maintaining individual autonomy,
making it particularly suitable for distributed energy management systems.
However, applying multi-agent RL to building energy management still faces three critical challenges:

\begin{figure}[!t]
  \centering
  \includegraphics[width=\linewidth]{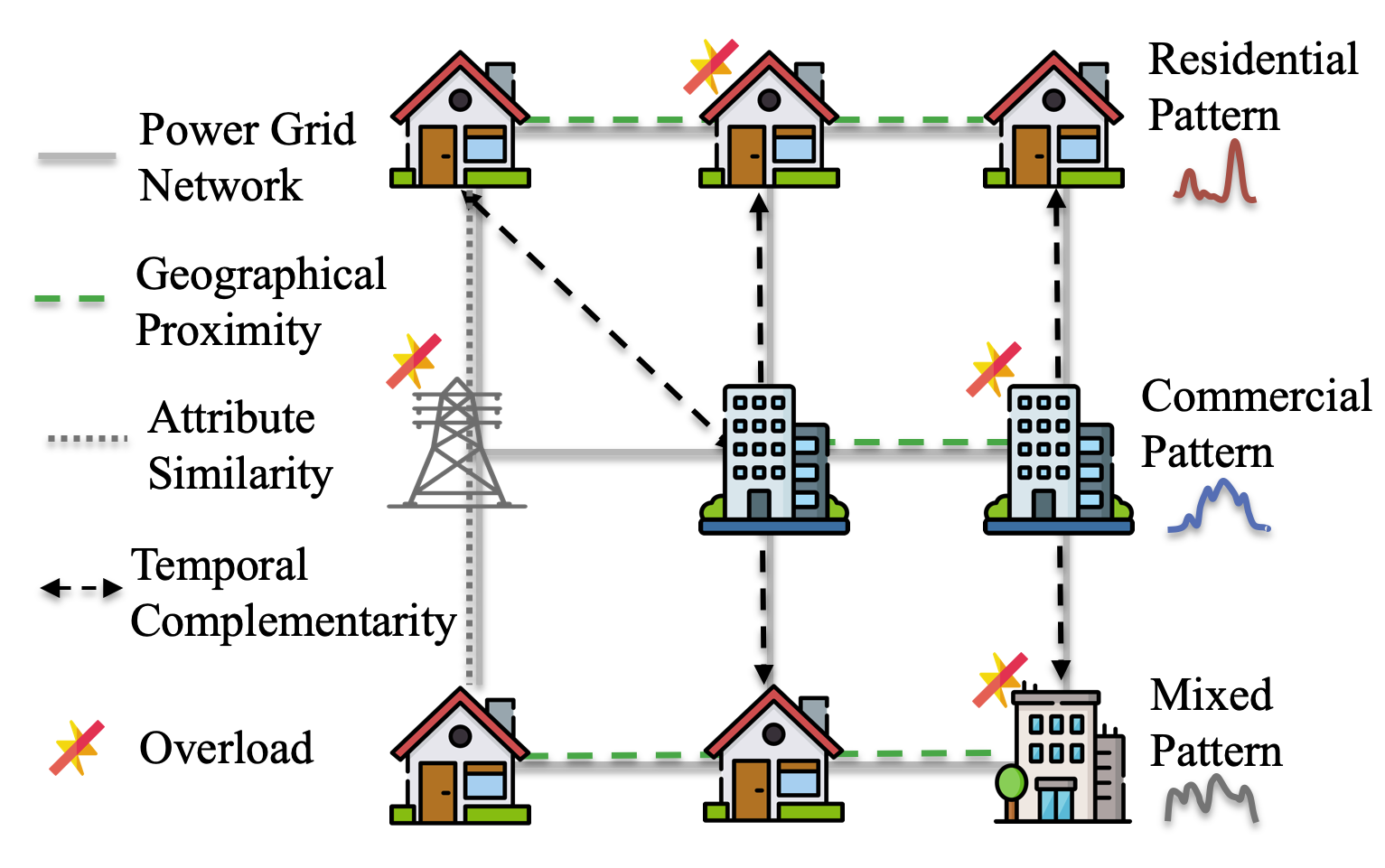}
  \caption{Complex spatial-temporal relationships and energy consumption patterns in multi-building systems: challenges of uncoordinated control and potential safety violations.}
  \label{fig:intro}
\end{figure}


\begin{itemize}
\item \textbf{Insufficient spatial-temporal information exploitation:} 
Current approaches fail to effectively exploit spatial-temporal information in building energy systems~\cite{haidar2023selective,chen2023graph}.
As shown in Figure~\ref{fig:intro}, buildings exhibit multiple types of relationships alongside their distinct energy consumption patterns.
The spatial relationships include geographical proximity, attribute similarity, and temporal complementarity among different building types.
Buildings close to each other will have heat transfer and shading effects among neighboring structures, residential buildings show double-peak patterns with morning and evening usage spikes, while commercial buildings maintain relatively stable consumption during business hours. 
However, existing methods ignore these hidden dependencies and patterns that are crucial for coordinated decision-making.
Without proper spatial-temporal feature extraction and modeling mechanisms, 
buildings make decisions based on limited local observations, leading to suboptimal performance.
For example, during summer peak hours, all buildings independently decide to pre-cool their spaces before peak pricing begins.
This simultaneous cooling demand overloads the electrical grid, 
causing supply instability and higher costs for everyone.

\item \textbf{Lack of rigorous safety guarantees:} 
Existing control methods lack rigorous safety guarantees~\cite{wang2025safe,ding2022safe} or rely on soft constraints, such as penalty terms in RL algorithms, 
to discourage unsafe behaviors.
However, these methods cannot provide mathematical guarantees against constraint violations~\cite{brunke2022safe}.
During the learning and exploration process, control systems may still take unsafe actions, such as exceeding battery capacity limits or causing equipment overloads.
Moreover, when buildings operate as isolated islands with limited local observations and broken information connections, 
they are more likely to exhibit safety warnings and uncoordinated consumption patterns, leading to power grid-level equipment damage risks and  overloads,
resulting in significant economic losses.

\item \textbf{System complexity and algorithmic challenges:} 
The inherent complexity of building energy systems poses significant algorithmic challenges for coordination~\cite{meng2024online,guo2022real}.
As illustrated in Figure~\ref{fig:intro}, the diverse building types and their complex interaction patterns create a challenging optimization landscape.
Multi-objective optimization requirements involving comfort, efficiency, and safety lead to complex trade-offs 
that are difficult to balance simultaneously~\cite{Jang_Yan_Spangher_Spanos_2024}.
The rapid development of distributed storage and renewable energy systems creates dynamic and uncertain environments 
with intermittent generation and variable demand patterns.
These factors, combined with the instability of RL algorithms in high-dimensional continuous spaces, 
make convergence slow and solution quality unpredictable~\cite{MARLISA,MA-HVAC}.
\end{itemize}

To address these multi-agent RL challenges in building energy management,
this paper proposes Spatial-Temporal Enhanced Safe Multi-Agent Coordination (STEMS),
a novel spatial-temporal aware safe multi-agent reinforcement learning framework for coordinated building energy management.
Our approach consists of two key components that work together to overcome the identified limitations.
First, we develop a spatial-temporal graph representation learning framework that effectively captures the complex relationships among buildings.
We propose a GCN-Transformer fusion architecture that processes both spatial dependencies through graph convolution 
and temporal patterns through attention mechanisms.
This framework includes a selective information construction mechanism that enables intelligent collaboration among buildings 
by dynamically determining what information to share based on spatial-temporal contexts.
The approach effectively handles high-dimensional state spaces through structured feature extraction and dimensionality reduction.
Second, we design a safety-constrained multi-agent RL algorithm that provides mathematical guarantees for safe operation.
We integrate Control Barrier Functions (CBFs) \cite{brunke2022safe} as safety shields that ensure control actions remain within safe operating bounds.
The algorithm incorporates a predictive safety checking mechanism that prevents unsafe actions from being executed 
while maintaining system performance.
This enables precise safety control in continuous action spaces, 
addressing the critical need for reliable operation in real-world building energy systems.
In summary, the main contributions of this work are threefold:
\begin{itemize}
\item 
We design a novel GCN-Transformer fusion module that effectively processes spatial-temporal dependencies 
with shared spatial and temporal information among buildings to improve the situation awareness 
and decision-making in building energy management systems.


\item 
We propose a constrained multi-agent RL framework that integrates CBFs
to provide mathematical guarantees for safe operation in building energy management.

\item 
We conduct extensive experiments on real-world datasets that validate the integration of spatial-temporal coordination and safety constraints enables \Name to achieve 21\% cost reduction over traditional methods, 65\% safety violation reduction compared with learning-based methods, and maintains the lowest discomfort rate, demonstrating simultaneous optimization of energy efficiency, operational safety, and occupant comfort.

\end{itemize}

The remainder of this paper is organized as follows.
Section II presents the system model and methodology, including problem formulation, the spatial-temporal graph representation learning framework, and the safety-constrained multi-agent RL algorithm.
Section III presents experimental settings and comprehensive results on real-world building energy datasets.
Section IV reviews related work, and Section V concludes the paper and discusses future research directions.

\section{System Model and Methodology}
\label{sec:Methodology}
\subsection{Problem Formulation}
\begin{figure*}
    \centering
    \includegraphics[width=0.8\linewidth]{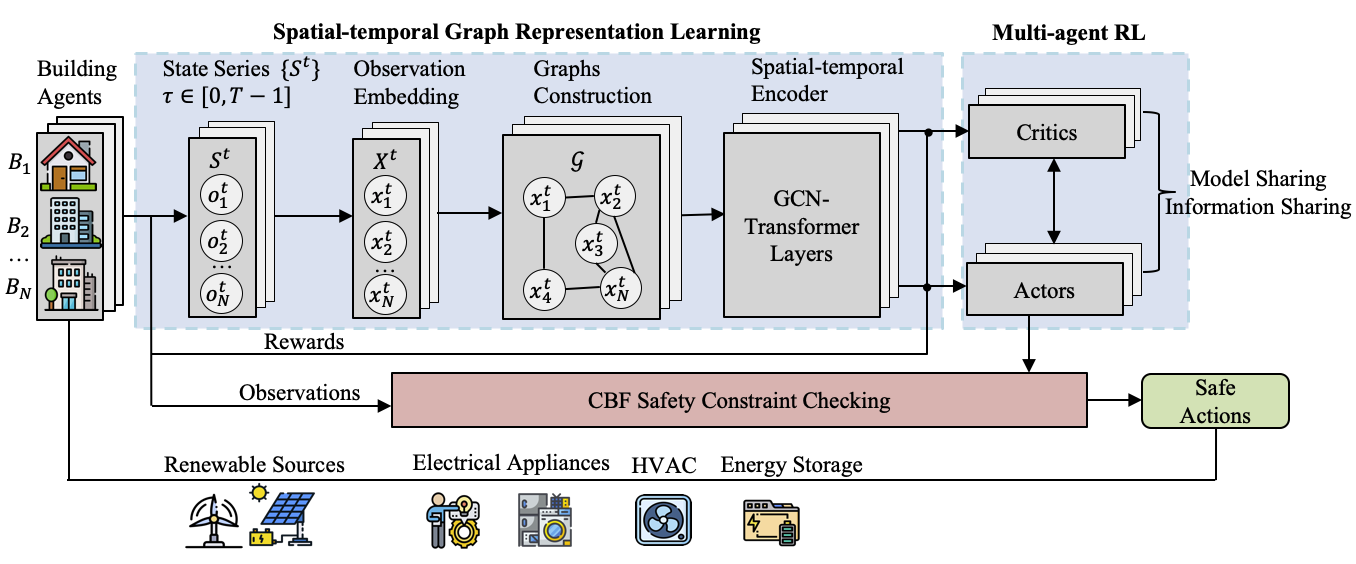}
    \caption{\Name Framework. Building agents collect state observations, which are processed through spatial-temporal graph learning to extract inter-building relationships and temporal patterns. The multi-agent RL module generates coordinated control actions, which are verified by CBF safety constraints before execution, with resulting rewards guiding the learning of RL.}
    \label{fig:framework_background}
    \end{figure*}
    
We consider a multi-building energy management system composed of $N$ buildings, 
denoted as the building set $\mathcal{B} = \{B_1, B_2, ..., B_N\}$. 
Each building is equipped with smart sensors and communication devices. 
This enables real-time monitoring of operational states and information sharing with neighboring buildings. 
The system includes diverse building types: residential, commercial, and mixed-use facilities. 
These buildings are distributed across geographical space with potential physical coupling relationships.


The inter-building relationships are represented by a communication graph 
$\mathcal{G} = (\mathcal{N}, \mathcal{E})$, 
where $\mathcal{N}$ denotes the node set (buildings) and $\mathcal{E}$ represents the edge set (connection relationships). 
Connection relationships between buildings are determined through two primary approaches: 
geographical proximity calculation based on Euclidean distance, 
and attribute similarity calculation based on building characteristics and usage patterns. 
The neighbor set of building $B_i$ is denoted as $\mathcal{N}_i$, 
representing buildings that can engage in information exchange.

The multi-building energy management challenge is formulated as a constrained multi-agent RL problem,
defined by the tuple:
\begin{equation}
\mathcal{M} = (\mathcal{S}, \mathcal{A}, P, \mathcal{R}, \mathcal{C}, \mathcal{G}, \gamma)
\end{equation}
where 
$\mathcal{R} = \{R_i\}$ denotes the reward function set, 
$\mathcal{C} = \{C_i\}$ represents the constraint function set, 
and $\gamma$ is the discount factor.

Each building agent observes a state space $\mathcal{S}_i = \{o_i, o_{j \in \mathcal{N}_i}\}$ 
that combines self-observations $o_i$ 
(including internal device states, environmental conditions, and economic signals) 
with shared information $o_{j \in \mathcal{N}_i}$ from neighboring buildings. 
The action space $\mathcal{A}_i$ are continuous control decisions 
for energy storage systems, Heating, Ventilation, and Air Conditioning (HVAC) operations, and other controllable devices.
The system aims to optimize long-term performance through reward functions $R_i$ 
that balance economic efficiency and operational stability, 
while satisfying safety constraints $C_i$ that ensure reliable and secure operation.

The optimization objective for each building agent is to learn an optimal policy 
$\pi_i^*: \mathcal{S}_i \mapsto \mathcal{A}_i$ that maximizes expected cumulative rewards:
\begin{equation}
\pi_i^* = \arg\max_{\pi_i} \mathbb{E}\left[\sum_{t=0}^T \gamma^t R_t^i \mid \pi_i\right]
\end{equation}
subject to safety constraints: $\mathbb{E}[C_i(s_t, a_t)] \geq 0, \forall t$
The safety constraints $C_i(s_t, a_t)$ include battery safety limits, 
power capacity constraints, and grid stability conditions, which are detailed in Section \ref{sec:safety-constrained}.

\subsection{Overall Framework}

To simultaneously address spatial-temporal dependency modeling, safety constraint guarantees, and multi-objective optimization challenges in building energy management, \Name adopts a modular design that integrates the spatial-temporal graph representation learning module, the CBF safety constraint checking module, and the multi-agent RL module.
Figure~\ref{fig:framework_background} illustrates the overall architecture and information flow among these components.
Initially, building agents collect comprehensive state observations $\mathcal{S}$ including internal device status, environmental conditions, and neighboring building information.
These observations are then processed by the spatial-temporal graph representation learning module through observation embeddings, graph convolution and temporal attention layers to extract spatial-temporal features that capture inter-building relationships and temporal patterns.
Subsequently multi-agent RL utilizes these enhanced spatial-temporal features to generate control actions $\mathcal{A}$ for each building through coordinated decision-making, where the rewards $\mathcal{R}$ and state transitions guide the continuous learning process.

The CBF safety constraint checking module verifies that all proposed actions satisfy operational safety requirements.  Once safe actions are executed in the environment, the resulting performance is evaluated through a multi-objective reward function that balances four key objectives:

\begin{equation}
R_t^i = R_{economic}^i + R_{stability}^i + R_{comfort}^i + R_{renewable}^i
\end{equation}

For building $i$ at time step $t$, the net electricity consumption is:
\begin{equation}
e_t^i = b_t^i + P_{HVAC,t}^i + P_{batt,t}^i - p_t^i
\end{equation}
where $b_t^i$ represents non-shiftable loads, 
$P_{HVAC,t}^i$ denotes HVAC power demand, 
$P_{batt,t}^i$ indicates battery charging/discharging power, 
and $p_t^i$ represents renewable energy generation.

The economic efficiency component minimizes electricity costs:
\begin{equation}
R_{economic}^i = -\mu \cdot v_t \cdot e_t^i
\end{equation}

The stability component combines grid-level coordination and building-level smooth operation:
\begin{align}
R_{stability}^i &= \alpha_{grid} \cdot \left(1 - \frac{\sum_{j=1}^N \max(0, e_t^j)}{P_{grid,max}}\right)^2 \nonumber \\
&\quad + \alpha_{build} \cdot \left(1 - \frac{|e_t^i|}{P_{building,max}^i}\right) \\ 
& \quad - \beta_{ramp} \cdot \frac{|e_t^i - e_{t-1}^i|}{P_{building,max}^i}
\end{align}

The comfort component maintains indoor temperature close to user preferences:
\begin{equation}
R_{comfort}^i = -\lambda_{indoor} \cdot |T_t^{in,i} - T_{ref}^i|^2
\end{equation}

The renewable energy component maximizes renewable energy utilization:
\begin{equation}
R_{renewable}^i = \xi \cdot \min\left(\frac{p_t^i}{p_t^i + \max(0, e_t^i)}, 1\right)
\end{equation}

These four components work together to achieve balanced performance. 
The economic term minimizes electricity costs by optimizing consumption timing and renewable energy use.
For system stability, the grid-level coordination with smooth building operations helps buildings work together to avoid overloads while maintaining steady power consumption.
The comfort term allows flexible trade-offs between energy efficiency and occupant comfort based on real-time conditions and the renewable component maximizes green energy utilization across all buildings.

\subsection{Spatial-Temporal Graph Representation Learning Framework}

The spatial-temporal graph representation learning framework captures the complex interdependencies 
among buildings and their temporal evolution patterns to enhance decision-making capabilities.
This framework leverages graph neural networks to encode spatial relationships 
and temporal attention mechanisms to model dynamic patterns in energy consumption and environmental conditions.

\subsubsection{Graph Construction and Node Features}

The building interaction graph $\mathcal{G}_{build} = (\mathcal{V}, \mathcal{E}_{build})$ is constructed 
where each node represents a building and edges encode interaction relationships.
Node features for building $i$ at time $t$ include:
\begin{equation}
\mathbf{x}_i^t = [\mathbf{s}_i^t, \mathbf{e}_i^t, \mathbf{c}_i^t]
\end{equation}
where $\mathbf{s}_i^t$ represents building state features (State of Charge (SOC), indoor temperature, device status),
$\mathbf{e}_i^t$ denotes environmental features (outdoor temperature, solar irradiance, electricity price),
and $\mathbf{c}_i^t$ captures building characteristics (type, capacity, efficiency parameters).

Edge weights are computed based on the connection strength between buildings:
\begin{align}
w_{ij} &= \alpha \cdot \exp\left(-\frac{d_{ij}^2}{2\sigma_d^2}\right)  + \beta \cdot \exp\left(-\frac{\|\mathbf{f}_i - \mathbf{f}_j\|^2}{2\sigma_f^2}\right)
\end{align}
where $d_{ij}$ represents geographical distance, 
$\mathbf{f}_i$ and $\mathbf{f}_j$ are building attribute vectors,
and $\alpha, \beta$ are weighting parameters for geographical and attribute-based similarities.

\subsubsection{Spatial Feature Encoding}

Graph convolutional layers aggregate information from neighboring buildings 
to capture spatial dependencies in energy consumption patterns.
The spatial encoding for building $i$ at layer $l$ is computed as:
\begin{equation}
\mathbf{h}_i^{(l+1)} = \sigma\left(\mathbf{W}^{(l)} \sum_{j \in \mathcal{N}_i \cup \{i\}} \frac{w_{ij}}{\sqrt{d_i d_j}} \mathbf{h}_j^{(l)}\right)
\end{equation}
where $\mathbf{h}_i^{(l)}$ represents the hidden state of building $i$ at layer $l$,
$\mathbf{W}^{(l)}$ is the learnable weight matrix,
$d_i$ is the degree of node $i$, and $\sigma$ is the activation function.

\subsubsection{Temporal Pattern Modeling}

A multi-head attention mechanism captures temporal dependencies 
in building energy consumption and environmental patterns.
For building $i$, the temporal attention weights are computed as:
\begin{equation}
\alpha_{i,t,\tau} = \frac{\exp(\mathbf{q}_{i,t}^T \mathbf{k}_{i,\tau})}{\sum_{\tau'=t-T}^t \exp(\mathbf{q}_{i,t}^T \mathbf{k}_{i,\tau'})}
\end{equation}
where $\mathbf{q}_{i,t}$ and $\mathbf{k}_{i,\tau}$ are query and key vectors derived from building states,
and $T$ is the temporal window size.

The temporal representation is obtained through weighted aggregation:
\begin{equation}
\mathbf{z}_i^t = \sum_{\tau=t-T}^t \alpha_{i,t,\tau} \mathbf{v}_{i,\tau}
\end{equation}
where $\mathbf{v}_{i,\tau}$ represents the value vector at time $\tau$.

The final spatial-temporal representation combines spatial and temporal features:
\begin{equation}
\mathbf{r}_i^t = \mathbf{W}_s \mathbf{h}_i^{(L)} + \mathbf{W}_t \mathbf{z}_i^t + \mathbf{b}
\end{equation}
where $\mathbf{W}_s$ and $\mathbf{W}_t$ are learnable projection matrices, 
and $\mathbf{b}$ is the bias term.

\subsection{Safety-Constrained Decision Making}
\label{sec:safety-constrained}


Building energy systems require strict safety constraints to prevent equipment damage and ensure reliable operation. 
Battery overcharging can cause thermal runaway and fire hazards. 
Power overloads can damage electrical equipment and cause grid instability.
Our safety-constrained approach uses CBFs to provide mathematical safety guarantees under feasible conditions, unlike penalty-based methods that may still produce unsafe actions during training and exploration.

\subsubsection{Control Barrier Function Framework}

To enforce the safety constraints $c_i(s_t, a_t) \geq 0$ defined in the problem formulation, 
we construct corresponding CBFs that provide implementable constraint enforcement mechanisms. 
We define three key safety constraints for building energy management:

\textbf{Battery Safety:} Energy storage must operate within safe SOC limits to prevent thermal runaway and equipment damage:
\begin{align}
h_{battery}^i(s,a) &= \min(SOC_{max} - SOC_{t+1}^i, \nonumber \\
&\quad\quad\quad SOC_{t+1}^i - SOC_{min}) \geq 0
\end{align}
The battery state is updated by:
$SOC_{t+1}^i = SOC_t^i + \frac{a_{battery}^i \cdot \Delta t}{\text{Capacity}_i}$.

\textbf{Building Safety:} Building power consumption cannot exceed capacity to prevent equipment overload:
\begin{equation}
h_{power}^i(s,a) = P_{building,max}^i - |e_t^i| \geq 0
\end{equation}

\textbf{Grid Safety:} Total system load must stay within grid limits to maintain system stability:
\begin{equation}
h_{grid}(s,a) = P_{grid,max} - \sum_{i=1}^N \max(0, e_t^i) \geq 0
\end{equation}


For each continuous action $a_t^i \in \mathbb{R}^{|\mathcal{A}_i|}$ generated by the actor network, we conduct safety verification, which is detailed in Algorithm~\ref{alg:safety-action-selection}, ensuring that all building control actions satisfy safety constraints before execution.

First, we evaluate whether the proposed action directly satisfies all safety constraints:
\begin{equation}
h_j(s_t, a_t^i) \geq 0, \quad \forall j \in \{\text{battery, power, grid}\}
\end{equation}

If all constraints are satisfied, the action is executed directly: $a_t^{i,safe} = a_t^i$.

Then, if any constraint is violated, we solve the CBF-quadratic programming (QP) optimization problem to find the closest safe action:
\begin{align}
a_t^{i,safe} = \arg\min_{u} \frac{1}{2}\|u - a_t^i\|^2 \quad \text{s.t.} \quad h_j(s_t, u) \geq 0, \forall j
\end{align}
The safe action space is defined as:
\begin{equation}
\mathcal{A}_i^{safe} = \{a \in \mathcal{A}_i : \text{safety constraints satisfied}\}
\end{equation}
This ensures the executed action is as close as possible to the original policy output while maintaining safety.
If it is feasible, the action is safe. Otherwise, emergency conservative actions including stopping battery charging and reducing HVAC power will be applied. 
This approach ensures reliable building energy management operations through mathematical safety guarantees.

\subsection{Multi-Agent Training Algorithm}

The proposed safety-constrained multi-agent RL algorithm integrates 
the spatial-temporal graph representation learning framework with safety constraint enforcement 
to achieve coordinated and safe energy management across multiple buildings. The complete training procedure is detailed in Algorithm~\ref{alg:training-procedure}, which shows the interaction between spatial-temporal learning, safety verification, and policy optimization.

\textbf{Actor-Critic Architecture:}
Each building agent $i$ employs an actor-critic architecture where the actor network $\pi_\theta^i$ 
generates control actions and the critic network $V_\phi^i$ estimates state values.
The spatial-temporal representation $\mathbf{r}_i^t$ serves as input to both networks:
\begin{align}
a_t^i &= \pi_\theta^i(\mathbf{r}_i^t) + \epsilon_t \\
V_t^i &= V_\phi^i(\mathbf{r}_i^t)
\end{align}
where $\epsilon_t$ represents exploration noise and $\theta, \phi$ are network parameters.


\begin{algorithm}
    \caption{Multi-Building Safety Verification}
    \label{alg:safety-action-selection}
    \begin{algorithmic}[1]
    \Require Continuous actions $\{a_t^i\}_{i=1}^N$ from actor networks, current state $s_t$
    \Ensure Safe actions $\{a_t^{i,safe}\}_{i=1}^N$
    \For{each building $i = 1$ to $N$}
        \State Check if $h_j(s_t, a_t^i) \geq 0, \forall j \in \{\text{battery, power, grid}\}$
        \If{All constraints satisfied}
            \State $a_t^{i,safe} = a_t^i$
        \Else
            \State Solve CBF-QP: $\min_u \frac{1}{2}\|u - a_t^i\|^2$ s.t. $h_j(s_t, u) \geq 0, \forall j$
            \If{CBF-QP has feasible solution}
                \State $a_t^{i,safe} =$ optimal solution
            \Else
                \State Apply emergency actions based on constraint violation type
                \State $a_t^{i,safe} =$ emergency action
            \EndIf
        \EndIf
    \EndFor
    \State \Return $\{a_t^{i,safe}\}_{i=1}^N$
\end{algorithmic}
\end{algorithm}

\textbf{Policy Update:}
The actor network is updated using the policy gradient with safety-constrained rewards:
\begin{equation}
\nabla_{\theta^i} J(\theta^i) = \mathbb{E}_{s^i,a^i \sim \mathcal{A}^{safe}_i}\left[\nabla_{\theta^i} \log \pi_{\theta^i}(a^i|s^i) \cdot A^i(s^i,a^i)\right]
\end{equation}
where $A^i(s^i,a^i) = R^i + \gamma V_{\phi^i}(s'^i) - V_{\phi^i}(s^i)$ is the advantage function for agent $i$,
and the expectation is taken only over the safe action space.

\textbf{Critic Update:}
The critic network is updated to minimize the temporal difference error:
\begin{equation}
L(\phi^i) = \mathbb{E}_{s^i,a^i \sim \mathcal{A}^{safe}_i}\left[(R^i + \gamma V_{\phi^i}(s'^i) - V_{\phi^i}(s^i))^2\right]
\end{equation}

\textbf{Graph Representation Update:}
The spatial-temporal graph representation parameters are updated end-to-end 
through backpropagation from the actor-critic losses:
\begin{equation}
\nabla_\psi L_{total} = \nabla_\psi J(\theta) + \nabla_\psi L(\phi)
\end{equation}
where $\psi$ represents the parameters of the graph neural network components.

\begin{algorithm}
\caption{Safety-Constrained Multi-Agent Training}
\label{alg:training-procedure}
\begin{algorithmic}[1]
\Require Initial policy parameters $\{\theta^i\}_{i=1}^N$, critic parameters $\{\phi^i\}_{i=1}^N$, spatial-temporal graph parameters $\psi$
\Ensure Trained  policy, critic and spatial-temporal graph networks
\For{episode $e = 1$ to $E_{max}$}
    \State Initialize environment and building states $\{s_0^i\}_{i=1}^N$
    \For{time step $t = 0$ to $T-1$}
        \State Compute spatial-temporal representations $\{\mathbf{r}_t^i\}_{i=1}^N$ using graph networks
        \State Generate raw actions $\{a_t^i\}_{i=1}^N$ from actor networks $\{\pi_{\theta^i}\}_{i=1}^N$
        \State Apply Algorithm \ref{alg:safety-action-selection} to get safe actions $\{a_t^{i,safe}\}_{i=1}^N$
        \State Execute safe actions and observe rewards $\{R_t^i\}_{i=1}^N$ and next states $\{s_{t+1}^i\}_{i=1}^N$
        \State Store experience tuples $\{(s_t^i, a_t^{i,safe}, R_t^i, s_{t+1}^i)\}_{i=1}^N$
    \EndFor
    \State Compute advantage functions $\{A_t^i\}$ using critic networks
    \State Update actor networks: $\theta^i \leftarrow \theta^i + \alpha_\pi \nabla_{\theta^i} J(\theta^i)$
    \State Update critic networks: $\phi^i \leftarrow \phi^i + \alpha_v \nabla_{\phi^i} L(\phi^i)$
    \State Update spatial-temporal graph parameters: $\psi \leftarrow \psi + \alpha_g \nabla_\psi L_{total}$
\EndFor
\State \Return Trained policies $\{\pi_{\theta^i}^*\}_{i=1}^N$, critics $\{V_{\phi^i}\}_{i=1}^N$, and spatial-temporal graph networks $\psi$
\end{algorithmic}
\end{algorithm}

This integrated approach ensures that the learned policies not only optimize performance 
but also maintain strict adherence to safety requirements throughout the learning process.
The spatial-temporal graph representation enables effective coordination among buildings,
while the safety constraint enforcement provides reliable operation guarantees.

\section{Experiments}
\label{sec:exp}
In this section, we evaluate our proposed \Name method for coordinated building energy management.
We first present the experimental settings including simulation environment, baseline methods, and evaluation metrics. 
Then we analyze the experimental results to demonstrate the effectiveness of our approach.

\subsection{Experimental Setup}
\label{exp_set}

\begin{table*}[!ht]
  \centering
  \caption{Long-term Performance Comparison (12-Month Average Results). }
  \resizebox{0.8\textwidth}{!}{%
  \footnotesize
  \begin{tabular}{@{}l@{\hspace{6pt}}c@{\hspace{6pt}}c@{\hspace{6pt}}c@{\hspace{6pt}}c@{\hspace{6pt}}c@{\hspace{6pt}}c@{\hspace{6pt}}c@{}}
  \hline
  Method & Cost & Emission & Avg. Daily & Electricity & Ramping & Discomfort & Safety \\
  &  &  & Peak & Consumption & Rate & Rate & Viol. Rate \\
  \hline
  Rule-Based & 1.000 & 1.000 & 1.000 & 1.000 & 1.000 & 0.130 & 0.351 \\
  MPC & 0.872 & 0.914 & 0.983 & 0.975 & 0.981 & 0.654 & 0.330 \\
  Single-Agent SAC & 0.824 & 0.867& 0.856 & 0.813 & 0.925 & 0.485 & 0.223 \\
  MADDPG & 0.795 & 0.817 & 0.854 & 0.856 & 0.936 & 0.357 & 0.197 \\
  MetaEMS & 0.836 & \textbf{0.804} & 0.884 & 0.835 & 0.907 & 0.396 & 0.231 \\
  MARLISA & 0.847 & 0.875 & 0.826 & 0.838 & 0.956 & 0.237 & 0.214 \\
  MADCQ & 0.814 & 0.837 & 0.862 & 0.823 & 0.906 & 0.325 & 0.155 \\
  D-MAPPO & 0.803 & 0.834 & 0.841 & 0.816 & 0.912 & 0.152 & 0.198 \\
  \hline
  \textbf{\Name} & \textbf{0.792} & 0.824 & \textbf{0.821} & \textbf{0.805} & \textbf{0.883} & \textbf{0.132} & \textbf{0.056} \\
  \hline
  \end{tabular}
  }
  \label{tab:comprehensive_results}
\end{table*}
  
\subsubsection{Simulation Environment}

We conduct experiments based on the CityLearn environment\cite{citylearn}, 
a multi-building energy management simulation platform that provides realistic building energy dynamics and grid interactions.
The experimental dataset is derived from real building energy consumption data in Travis County, Texas. 
The dataset spans from August 2018 to August 2019, capturing a full year of operation under typical subtropical humid climate conditions: 
annual average temperature of 20.7°C, temperature range from -7.8°C to 42.8°C.
This climate profile creates significant cooling-dominated energy demands during summer months 
and moderate heating requirements in winter, providing realistic conditions for evaluating 
multi-building energy coordination strategies.
The data includes 1 hour interval measurements of electricity consumption, 
solar generation, battery storage, and environmental conditions. 
The experimental setup consists of 8 buildings in a neighborhood configuration 
with typical regional distribution (5 residential, 2 commercial, 1 mixed-use).
The 8 buildings are randomly selected from the Travis County dataset  based on above ratio to ensure 
representative diversity in building characteristics and energy consumption patterns.
Most residential buildings  show dual-peak patterns with morning (6-9h) and evening (18-22h) usage spikes, 
reflecting typical household activities with strong seasonal variations. 
Commercial buildings demonstrate business-hour concentrated loads (9-17h) with midday peaks around 12-13h. 
Mixed-use buildings combine both residential and commercial characteristics with variable all-day consumption patterns. 
Each building is equipped with controllable devices including battery energy storage systems (BESS),  HVAC systems, solar photovoltaic panels, hot water tank.
Buildings can communicate with each other to share state information and coordinate energy management decisions. 



\subsubsection{Baseline Methods}

We compare our method (\Name) against the following baseline methods:

\textbf{Rule-Based}: Traditional rule-based control using time-of-use scheduling and temperature setpoint control.

\textbf{MPC}: Model Predictive Control using linear building thermal models and quadratic cost functions.

\textbf{Single-Agent SAC}: Independent Soft Actor-Critic reinforcement learning for each building without coordination\cite{sac}.

\textbf{MADDPG}: Multi-agent deep deterministic policy gradient with centralized training and decentralized execution\cite{lowe2017multi-maddpg}.

\textbf{MetaEMS}: Meta-learning based energy management system using model-agnostic meta-learning (MAML) framework\cite{MetaEMS}.

\textbf{MARLISA}: Multi-agent reinforcement learning with independent soft actor-critic operating independently\cite{MARLISA}.

\textbf{MADCQ}: Multi-agent deep constrained Q-learning with independent building-level optimization and heuristic constraint handling\cite{saberi2024madcq}.

\textbf{D-MAPPO}: Distributed multi-agent proximal policy optimization with soft penalty constraints for improved training stability\cite{tariq2025dmappo}.
\subsubsection{Evaluation Metrics}

We evaluate methods using the following key performance metrics:

\textbf{Cost}: Total electricity cost including energy charges and demand charges. 

\textbf{Emission}: Carbon emissions from electricity consumption based on grid carbon intensity.

\textbf{Avg. Daily Peak}: Average of maximum daily power demand across all buildings:
$$P_{avg} = \frac{1}{D} \sum_{d=1}^{D} \max_{t \in d} \sum_{i=1}^{N} e_{i,t}$$
where $D$ is the number of days and $N$ is the number of buildings.

\textbf{Electricity Consumption}: Total electricity consumption from the grid.

\textbf{Ramping Rate}: Grid load change rate measuring power fluctuations:
$$R = \frac{1}{T-1} \sum_{t=2}^{T} |e_t - e_{t-1}|$$
where $e_t$ is the total grid load at time $t$.

\textbf{Discomfort Rate}: Proportion of occupied timesteps when indoor temperature deviates from setpoint by more than 2°C:
$$D_{rate} = \frac{1}{T_{occ}} \sum_{t=1}^{T} \sum_{i=1}^{N} \mathbf{1}(O_{i,t} > 0) \cdot \mathbf{1}(|T_{i,t} - T_{i,t}^{set}| > 2)$$
where $T_{occ}$ is the total number of occupied timesteps across all buildings, $T_{i,t}^{set}$ is the setpoint temperature of building $i$ at time $t$, and $O_{i,t}$ indicates occupancy status.

\textbf{Safety Viol. Rate}: Proportion of timesteps violating safety constraints.

All metrics except Discomfort Rate and Safety Violations Rate are normalized relative to the rule-based baseline performance to enable fair comparison across different methods and scenarios.

\begin{figure*}[!h]
  \centering
  \includegraphics[width=0.9\linewidth]{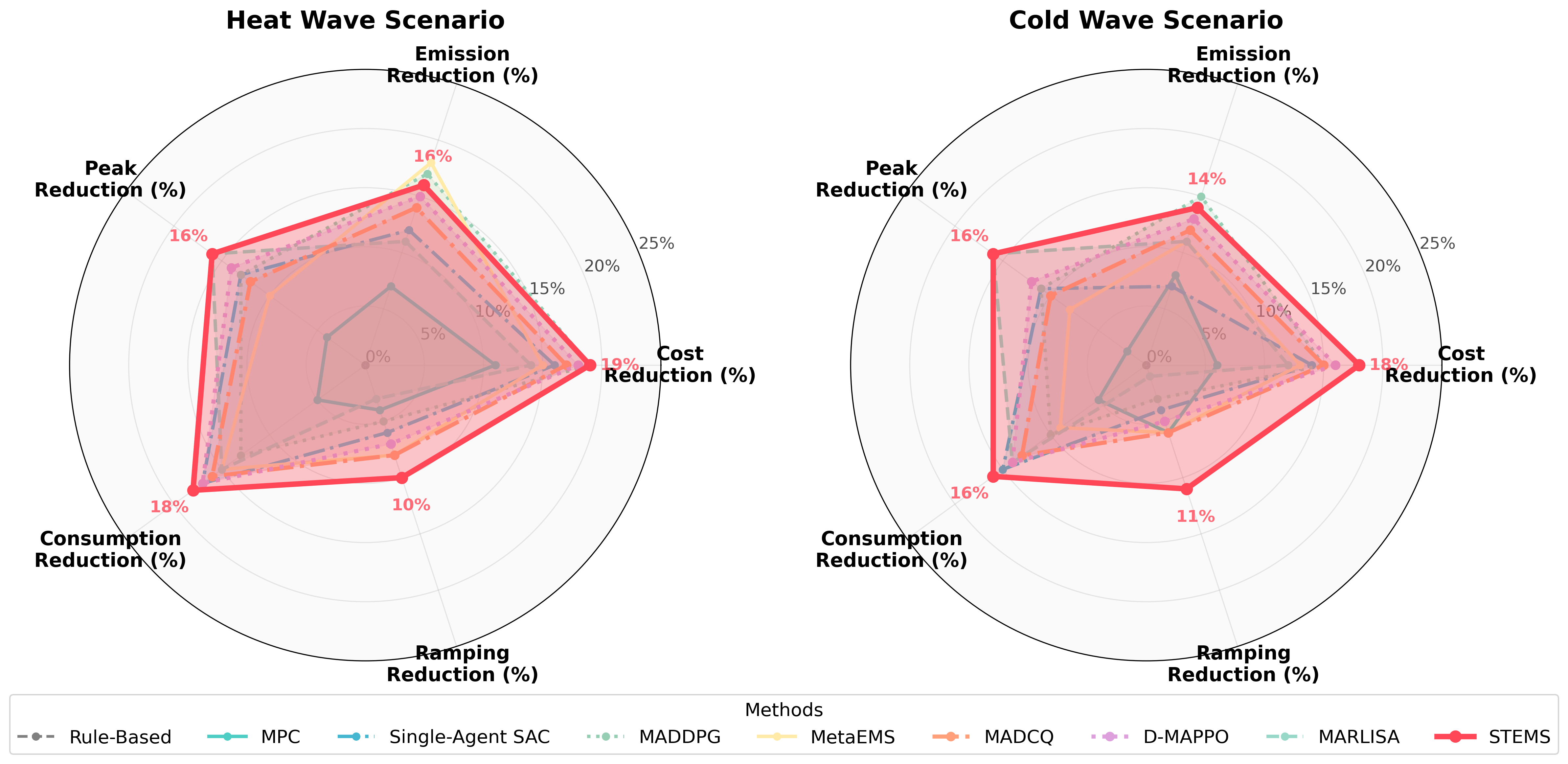}
  \caption{Performance comparison between Heat Wave and Cold Wave scenarios across five key metrics: 
  Cost Reduction, Emission Reduction, Peak Reduction, Consumption Reduction, and Ramping Reduction. 
  \Name (red) achieves 18-19\% cost reduction and 16\% peak reduction in both scenarios. 
  }
  \label{fig:extreme_weather_radar}
  \end{figure*}

\subsubsection{Implementation Details}

All algorithms are implemented in Python using PyTorch. 
The training dataset contains 8760 time steps covering a full year of operation with 1-hour intervals.
The graph neural network uses 3 layers with 64 hidden units each. 
The temporal attention mechanism has 4 heads with 32-dimensional embeddings. 
Actor and critic networks have 2 hidden layers with 128 units each.
The weighting parameters are: $\mu = 1.0$ (economic baseline), $\alpha_{grid} = 0.5$ (grid stability), 
$\alpha_{build} = 0.3$ (building stability), $\beta_{ramp} = 0.2$ (power ramp penalty), 
$\lambda_{indoor} = 0.4$ (indoor comfort), and $\xi = 0.6$ (renewable energy).
Training uses the Adam optimizer with learning rate $3 \times 10^{-4}$. 
The discount factor $\gamma$ is set to 0.99.
The model converges within 15 epochs, with each epoch taking approximately 3 minutes.
Total training time is approximately 45 minutes on an NVIDIA RTX 5090 GPU.
All baseline methods are implemented following their original paper parameters and fine-tuned to achieve optimal performance in our experimental setting.

\subsection{Performance Evaluation}

We evaluate the learned policies after training completion. All reported 
metrics are computed using the converged policies during evaluation episodes, 
averaged over 5 random seeds to ensure statistical reliability.
Table~\ref{tab:comprehensive_results} presents the long-term performance comparison based on 12 months of continuous operation data.
 The results represent averaged performance across different seasons and weather conditions, demonstrating the sustained effectiveness of \Name in real-world deployment.

Over the 12-month evaluation period, \Name shows consistent improvement across almost all metrics compared to baseline methods.
\Name achieves 0.792 cost, 0.821 peak, 0.883 ramping rate, 0.132 discomfort rate, and 0.056 safety violations rate, demonstrating 1-2.5\% cost reduction and over 65\% safety improvement over state-of-the-art baselines.
Traditional control approaches exhibit significant limitations. 
Rule-based control suffers from rigid scheduling that cannot adapt to dynamic conditions, resulting in 0.351 safety violations rate. 
MPC relies on simplified linear models, achieving only modest improvements with 0.330 violations and 0.654 discomfort rate.
For learning-based baselines, Single-Agent SAC shows competitive economic performance with 0.824 cost but suffers from lack of coordination.
MADDPG and MetaEMS incorporate multi-agent learning but lack explicit safety mechanisms, leading to 0.197 and 0.231 violations respectively. 
MARLISA achieves reasonable performance but still exhibits 0.214 violations due to absence of safety constraints.
MADCQ employs heuristic constraint handling, achieving 0.155 violations, better than other learning methods but still 2.8$\times$ higher than \Name. 
D-MAPPO uses soft penalty constraints, maintaining good comfort with 0.152 discomfort rate but exhibiting 0.198 violations as penalties cannot ensure hard constraints.
\Name's integrated approach delivers best overall performance through three key advantages. GCN-based spatial coordination enables explicit inter-building information sharing to reduce the cost and Transformer-based temporal planning captures daily and weekly patterns for smooth control with 0.883 ramping rate versus 0.906-0.956 for baselines.
Furthermore, CBF-based mathematical safety guarantees reduce violations to 0.056, representing 65\% improvement compared with recent baselines MADCQ and D-MAPPO, providing formal guarantees essential for real-world deployment.

We also conduct case studies during extreme weather periods. 
\textbf{Heat Wave} conditions where temperature $\geq$ 30°C, occurring 14.8\% of the year, create high cooling demand with enhanced solar generation. 
\textbf{Cold Wave} conditions where temperature $\leq$ 0°C, occurring 1.2\% of the year, require substantial heating with minimal solar availability. 
These scenarios test system performance when individual components (batteries, HVAC systems) operate under stress.

\begin{figure}[!h]
  \centering
  \includegraphics[width=1\linewidth]{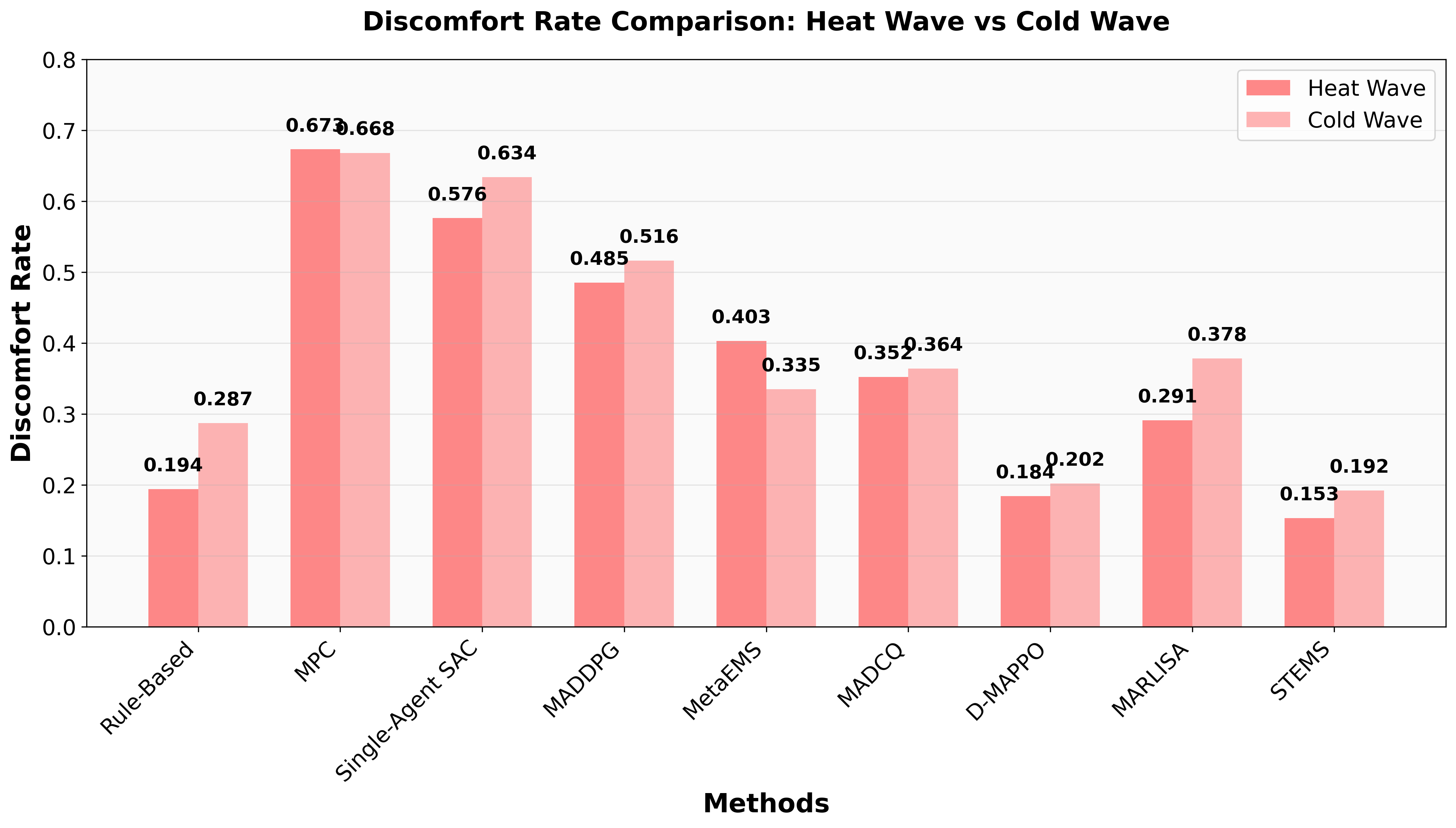}
  \caption{Discomfort rate comparison between Heat Wave and Cold Wave scenarios. 
  \Name maintains the lowest discomfort rate  across all evaluated methods. 
  }
  \label{fig:extreme_weather_discomfort}
  \end{figure}

\begin{figure}[!h]
  \centering
  \includegraphics[width=1\linewidth]{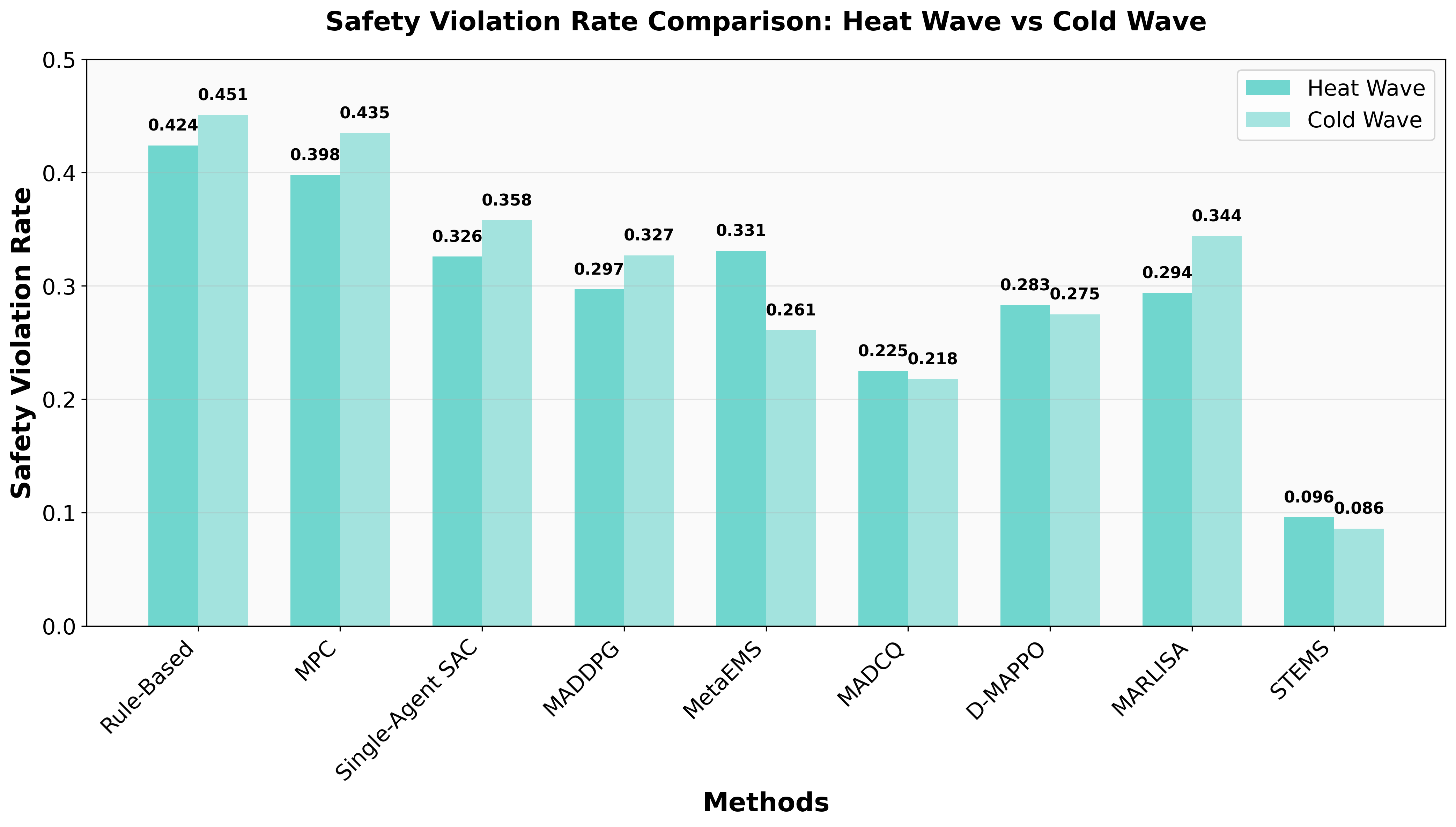}
  \caption{Safety violation rates comparison between Heat Wave and Cold Wave scenarios. 
  \Name achieves the lowest safety violation rates across all evaluated methods.
  }
  \label{fig:extreme_weather_safety}
  \end{figure}

Figure~\ref{fig:extreme_weather_radar} presents performance analysis under extreme weather conditions. 
The radar charts display improvement rates relative to Rule-Based baseline across five key metrics, providing intuitive comparison across multiple dimensions. 
Under Heat Wave conditions, \Name achieves 19\% cost reduction, 16\% emission reduction, 16\% peak reduction, 18\% consumption reduction, and 10\% ramping reduction. 
The Cold Wave scenario demonstrates similar performance with 18\% cost reduction, 14\% emission reduction, 16\% peak reduction, 16\% consumption reduction, and 11\% ramping reduction. 
The consistent radar patterns across both scenarios validate robust performance under contrasting thermal challenges.
Figure~\ref{fig:extreme_weather_discomfort} reveals distinct comfort management strategies. 
\Name maintains the lowest discomfort rates with 0.153 for Heat Wave and 0.192 for Cold Wave, closely matched by D-MAPPO with 0.184 and 0.202. 
The tight comfort control demonstrates effective thermal management even when HVAC systems operate near capacity limits during extreme temperatures.
Figure~\ref{fig:extreme_weather_safety} shows safety performance differences. 
\Name achieves the lowest violation rates with 0.096 for Heat Wave and 0.086 for Cold Wave, representing 77-81\% reduction compared to Rule-Based control. 
MADCQ achieves moderate safety with 0.225 and 0.218 through heuristic constraint handling, while D-MAPPO shows higher violations with 0.283 and 0.275 despite good comfort performance. 
The performance gap highlights that penalty-based approaches and heuristic methods cannot provide the formal safety guarantees achieved by CBF mathematical optimization, especially critical when systems operate under stress during extreme weather events.
\Name's spatial-temporal coordination enables effective energy management by optimizing battery 
reserves and coordinating inter-building information sharing before extreme events or during peak stress.
Additionally, CBF safety guarantees maintain constraint satisfaction even when systems operate at their limits during simultaneous peak demand.
In contrast, other baselines' constraint handling approaches fail to provide formal guarantees under stress.
  
\subsection{System Visualization}

\begin{figure}[!h]
\centering
\includegraphics[width=0.6\linewidth]{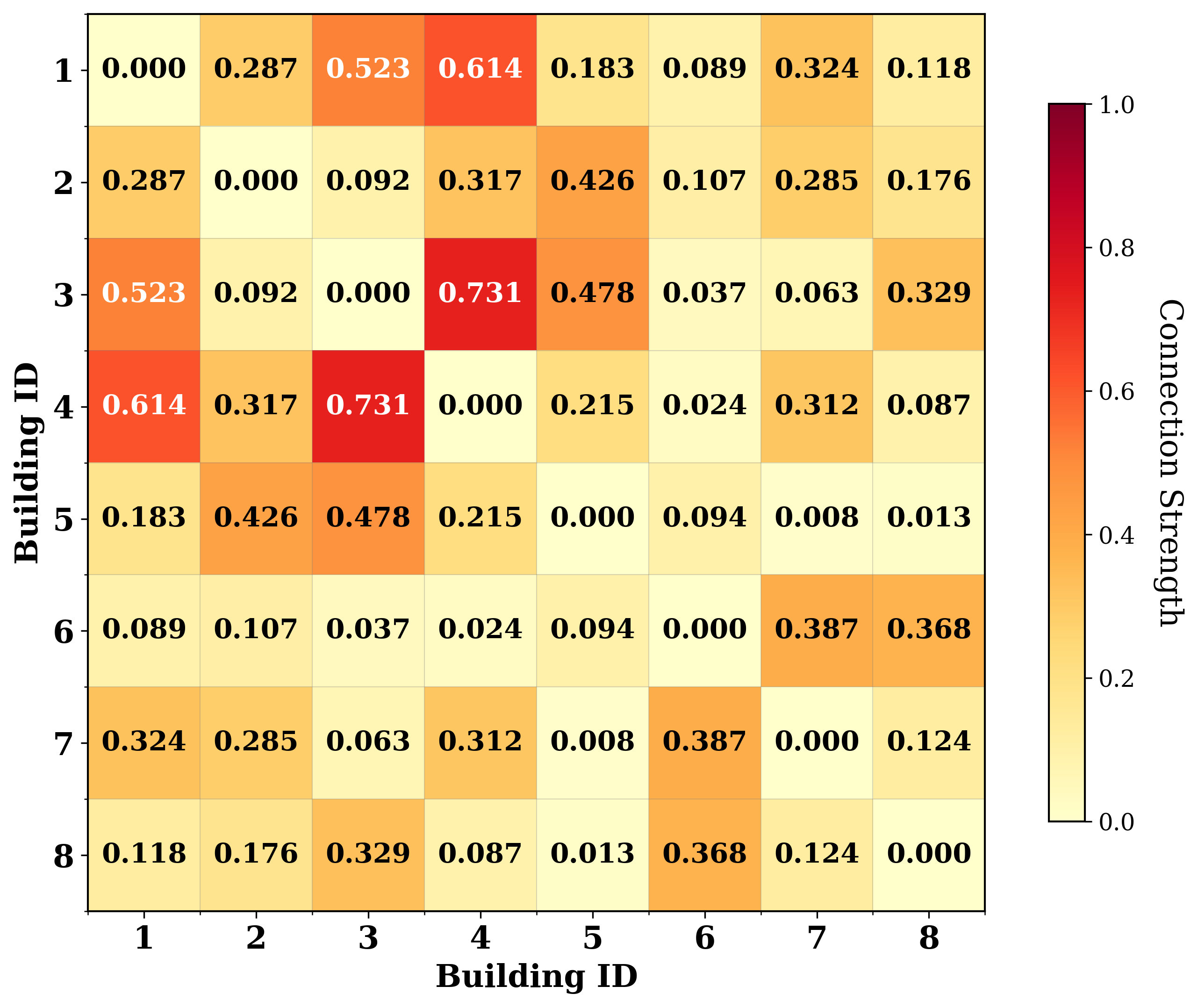}
\caption{Building connection weight heatmap showing learned spatial relationships among 8 buildings derived from geographic proximity and attribute similarity for coordinated energy management.}
\label{fig:spatial_temporal_core}
\end{figure}
To intuitively understand the spatial and temporal relationships learned by \Name, in this subsection, we inspect the learned building attention weight matrix and temporal attention.
We randomly select a set of buildings which includes 5 residential, 2 commercial and 1 mixed buildings . 
The building connection weight heatmap in Figure~\ref{fig:spatial_temporal_core} reveals the learned spatial relationships among 
the 8 buildings in the system. The visualization demonstrates a heterogeneous connection pattern with connection weights ranging from 0.008 to 0.731, reflecting \Name's ability to identify coordination opportunities and selective communication strategies. The strongest connections emerge between Buildings 3-4 with 0.731 and Buildings 1-4 with 0.614, indicating critical coordination pairs within the residential cluster that enable effective load balancing through information sharing. 
Residential buildings show varied connections with moderate to strong relationships, demonstrating selective coordination rather than uniform clustering. Commercial buildings exhibit moderate connection at 0.387, while the mixed building serves as a strategic coordination hub with balanced connections to both building types. Cross-type connections reveal strategic coordination patterns, with notable links such as Buildings 1-7 at 0.324 and 4-7 at 0.312 enabling load balancing during complementary demand periods. The learned connection weights automatically adapt to changing conditions, strengthening connections between buildings that benefit most from coordination while reducing unnecessary communication overhead.

\begin{figure}[!h]
  \centering
\includegraphics[width=\linewidth]{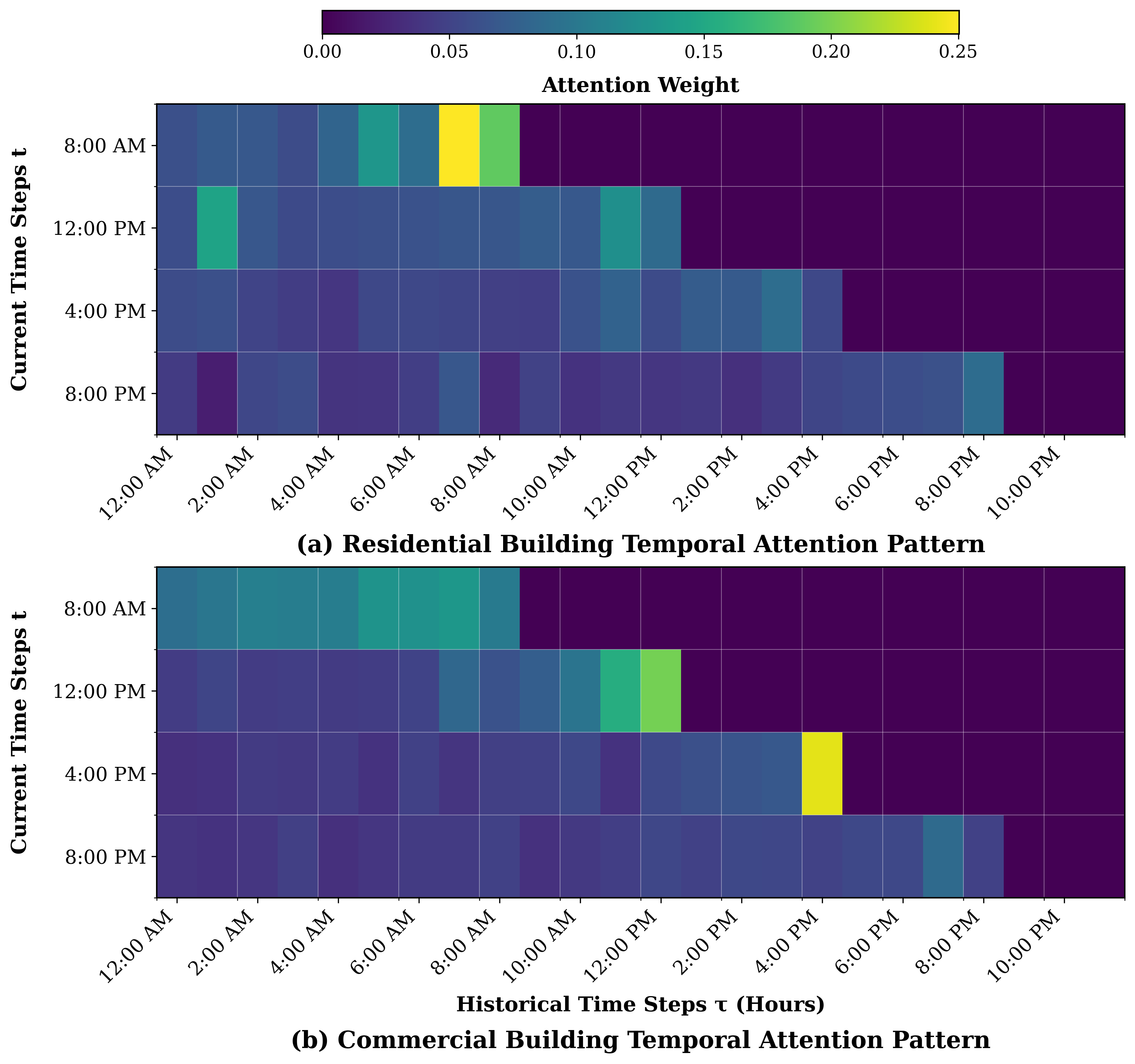}
\caption{Building-type-specific temporal attention patterns comparison: (a) Residential building showing distributed attention with evening and morning peaks, (b) Commercial building exhibiting concentrated work-hour attention with high regularity and minimal after-hours activity.}
\label{fig:building_attention_comparison}
\end{figure}

Then we inspect the temporal attention learned by temporal Transformer module. Figure~\ref{fig:building_attention_comparison} provides a detailed comparison of temporal attention patterns between residential and commercial buildings, demonstrating how \Name adapts its attention mechanisms to different building types' operational characteristics.
The residential building temporal attention pattern in Figure~\ref{fig:building_attention_comparison}a exhibits distinct characteristics reflecting typical household energy consumption behaviors. The visualization shows distributed attention during morning hours and peak attention at 8:00 PM, corresponding to residential energy consumption peaks when families engage in evening activities.
In contrast, the commercial building attention pattern in Figure~\ref{fig:building_attention_comparison}b demonstrates highly structured temporal dependencies aligned with business operations,
reflecting minimal energy activity after business hours when buildings operate in energy saving mode.

The complementary nature of these temporal patterns creates opportunities for coordinated load balancing through information sharing via the spatial connections. Residential buildings show peak activity during evening hours when commercial buildings are inactive. This integration demonstrates \Name's ability to learn and leverage both the physical similarities between buildings and their temporal operational patterns, enabling load shifting and peak shaving through the spatial network while respecting the natural characteristics of different building types.

\subsection{Robustness Analysis}
In this subsection, we evaluate the robustness of \Name through: 1) power outage scenarios to test resilience under unexpected grid failures, and 2) the cold-humid Vermont dataset consisting of same-type residential buildings to evaluate cross-climate generalizability.


\subsubsection{Power Outage Scenarios}

Power outages represent critical stress tests for building energy management systems, 
particularly during extreme weather when grid reliability decreases. We simulate stochastic 
power outage scenarios in CityLearn environment with configurations based on industry-standard reliability indices from U.S. EIA data: SAIFI=1.436 events/year 
and CAIDI=331.2 minutes/event (approximately 5.5-hour outages)~\cite{citylearn}. We evaluate system 
robustness under three weather conditions during which power outages occur: (1) \textbf{Year-round}: 
outages occur randomly throughout the year, (2) \textbf{Heat Wave} (T $\geq$ 30°C), and (3) \textbf{Cold Wave} (T $\leq$ 0°C). All 
methods were tested directly under outage conditions. Results averaged over 5 random seeds are presented in 
Table~\ref{tab:outage_robustness}. The unserved energy rate represents normalized energy 
demand unmet during outages (lower is better).
\begin{table*}[h]
\centering
\caption{Power Outage Robustness under Different Weather Conditions.}
\resizebox{0.8\textwidth}{!}{%
\scriptsize
\begin{tabular}{llccccccc}
\toprule
\textbf{Scenario} & \textbf{Metric} & \textbf{Rule-} & \textbf{Single-} & \textbf{MAD} & \textbf{Meta} & \textbf{MAD} & \textbf{D-} & \textbf{\Name} \\
& & \textbf{Based} & \textbf{Agent SAC} & \textbf{DPG} & \textbf{EMS} & \textbf{CQ} & \textbf{MAPPO} & \\
\midrule
\multirow{2}{*}{\textbf{Year-round}} 
& Unserved Energy Rate & 0.784 & 0.663 & 0.605 & 0.578  & 0.582& 0.554 & \textbf{0.553} \\
& Safety Viol. Rate & 0.412 & 0.283 & 0.254 & 0.291 & 0.214 & 0.253 & \textbf{0.082} \\
\midrule
\multirow{2}{*}{\textbf{Heat Wave}} 
& Unserved Energy Rate& 0.813 & 0.705 & 0.642 & 0.618 &  0.624 & 0.601 & \textbf{0.594} \\
& Safety Viol. Rate & 0.453 & 0.325 & 0.294 & 0.365 & 0.274 & 0.325 & \textbf{0.113} \\
\midrule
\multirow{2}{*}{\textbf{Cold Wave}} 
& Unserved Energy Rate& 0.842 & 0.736 & 0.674 & 0.648  & 0.653& 0.635 & \textbf{0.623} \\
& Safety Viol. Rate & 0.485 & 0.354 & 0.326 & 0.332 & 0.243 & 0.291 & \textbf{0.102} \\
\bottomrule
\end{tabular}%
}
\label{tab:outage_robustness}
\end{table*}

Table~\ref{tab:outage_robustness} demonstrates \Name's strong robustness across diverse 
weather conditions during power outages. In the year-round baseline scenario, \Name achieves 
0.553 unserved energy rate, outperforming almost all comparison methods through
GCN-based spatial coordination that enables load reduction strategies
across buildings during outages. Most significantly, \Name reduces safety violations to 
0.082, representing 62-80\% improvement over other methods, achieved through 
CBF-based mathematical safety guarantees that provide hard constraints under stress even without explicit power outage training.
Under extreme weather conditions, the performance gap becomes more pronounced. During Heat 
Wave scenarios with high cooling loads and abundant solar generation, \Name 
maintains 0.594 unserved energy rate and 0.113 safety violations. The Transformer's temporal 
planning recognizes outage risk during extreme weather and maintains higher battery 
reserves through conservative charging strategies, indirectly preparing for potential outages. 
In the most challenging Cold Wave scenarios with minimal solar generation and high 
heating demands, \Name achieves lowest 0.623 unserved energy rate and 0.102 safety violations. The GCN-based information sharing enables 
buildings to coordinate thermal management decisions based on neighbors' battery status and 
thermal conditions, optimizing cluster-wide energy utilization during prolonged outages.

Despite CBF protection, \Name maintains 0.082-0.113 residual safety violations during power 
outages. These occur during prolonged Cold Wave outages when battery capacity becomes 
insufficient to simultaneously maintain thermal comfort. In such 
cases, the CBF safety layer 
prioritizes hardware protection over temporary thermal discomfort, 
preventing catastrophic system failure.

\subsubsection{Cross-Climate Generalization}  
\begin{figure}[!h]
    \centering
    \includegraphics[width=0.95\linewidth]{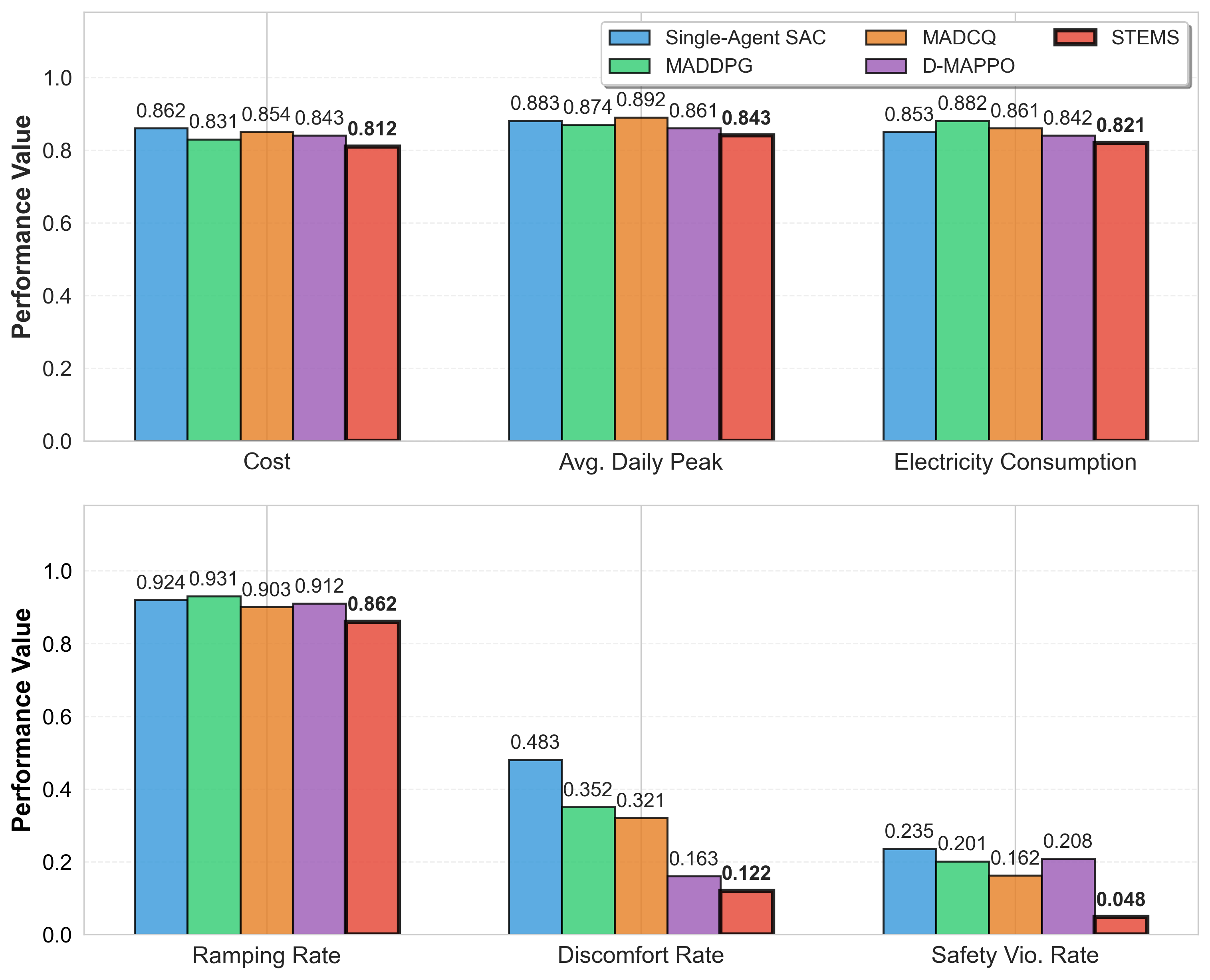}
    \caption{Vermont dataset performance comparison. \Name achieves the best overall 
    performance in cold-dominant climate.}
    \label{fig:vermont_combined}
\end{figure}

To evaluate \Name's generalization across different climate zones, we conduct experiments 
on the Vermont, Chittenden County dataset from CityLearn benchmark. This dataset exhibits 
cold-dominant characteristics with 23.6\% Cold Wave days (T $\leq$ 0°C) and only 1.8\% Heat Wave days (T $\geq$ 30°C). 
The temperature ranges from -28.3°C to 35.6°C with mean 9.1°C, representing a significantly 
different climate profile from the Texas dataset. The dataset contains 47 residential 
buildings with one year of hourly data (8,760 hours). We randomly selected 8 buildings and evaluated 12-month average results.
Figure~\ref{fig:vermont_combined} presents \Name's performance compared with four 
representative learning-based methods: Single-Agent SAC, MADDPG, and the two newly 
added baselines MADCQ and D-MAPPO. MetaEMS and MARLISA show similar performance 
patterns to MADDPG and are omitted for clarity.

From Figure~\ref{fig:vermont_combined}, \Name achieves 0.812 cost, 0.843 peak, and 
0.821 consumption, outperforming the second-best MADDPG by 2.4\%. For operational quality, 
\Name achieves 0.862 ramping rate and 0.122 discomfort rate. Most importantly, \Name reduces 
safety violations to 0.048, representing 70\% improvement over MADCQ and 77\% over D-MAPPO, 
validating that CBF-based guarantees remain effective across different climate zones.

Comparing Vermont with Texas reveals adaptation patterns. Energy efficiency shows 2-3\% 
higher cost in Vermont due to reduced load diversity. The Vermont dataset consists 
entirely of residential buildings with similar occupancy patterns, resulting in 
synchronized peak demands that limit peak shaving opportunities through temporal 
coordination. In contrast, Texas's mixed commercial-residential portfolio provides 
complementary load profiles where GCN-based information sharing enables more effective 
coordination. However, ramping rate improves to 0.862 as the Transformer better learns 
consistent residential consumption patterns. Safety violations decrease to 0.048 since 
homogeneous buildings share uniform constraints, simplifying CBF optimization. 
Discomfort rate improves to 0.122 through more uniform comfort-energy trade-offs.
These results validate \Name's cross-climate generalization capability and its adaptability 
to both heterogeneous and homogeneous building settings.


\subsection{Computational Complexity and Scalability}

We then report the computational efficiency and scalability of \Name, where Table~\ref{tab:inference_comparison} compares \Name with MPC and the state-of-the-art D-MAPPO and presents \Name's results across 5 to 15 buildings. For 8 buildings, \Name requires 45 minutes of offline training, comparable to D-MAPPO's 41 minutes, while enabling months of continuous operation. \Name's online inference completes in 24 ms, well within the typical 1-hour control cycle for real-time building energy management. In contrast, MPC requires 2.3 seconds per decision due to QP solving, introducing potential latency in time-critical scenarios.
Furthermore, we find that \Name achieves better safety at matched computational cost. At identical 8-building scale with similar training time, \Name achieves 0.056 safety violation rate versus D-MAPPO's 0.198, representing a 72\% reduction. This improvement stems from CBF-QP optimization that enforces hard constraints at each control step, adding less than 1 ms to the inference time. The result demonstrates that formal safety guarantees need not compromise computational efficiency.

We also evaluated scalability of \Name across 5 to 15 buildings, representing realistic deployment scenarios such as university campuses, corporate complexes, and residential communities \cite{tariq2025dmappo,safeMBRL,chen2024green}. Model size scales linearly with building count, from 1.51 MB to 3.22 MB, as shared spatial-temporal graph components remain constant while independent actor-critic networks grow proportionally. Training time follows $O(N^2)$ complexity due to graph message passing, increasing from 28 to 158 minutes. Despite this quadratic scaling, convergence times remain acceptable for offline training, while inference maintains $O(1)$ complexity per building through fixed network forward passes.


\begin{table}[h]
\centering
\caption{Computational Efficiency and Safety Performance Comparison. }
\resizebox{\columnwidth}{!}{%
\footnotesize
\begin{tabular}{lccccc}
\toprule
Method & Building & Training & Inference & Model & Safety \\
       & Count    & Time     & Time & Size  & Viol. Rate \\
\midrule
MPC    & 8        & N/A      & 2.3 s     & N/A   & 0.351 \\
D-MAPPO& 8        & 41 min  & 22 ms    & 1.18 MB & 0.198 \\
\midrule
\multirow{3}{*}{\Name}  & 5        & 28 min   & 17 ms    & 1.51 MB & 0.055 \\
                        & 8        & 45 min   & 24 ms     & 1.91 MB & 0.056 \\
                        & 15       & 158 min  & 30 ms    & 3.22 MB & 0.061 \\
\bottomrule
\end{tabular}%
}
\label{tab:inference_comparison}
\end{table}
\subsection{Ablation Study}

Table~\ref{tab:ablation} analyzes the contribution of each component in the \Name framework.

\begin{table*}[!t]
\centering
\caption{Ablation Study Results}
\resizebox{0.8\textwidth}{!}{%
\footnotesize
\begin{tabular}{@{}l@{\hspace{6pt}}c@{\hspace{6pt}}c@{\hspace{6pt}}c@{\hspace{6pt}}c@{\hspace{6pt}}c@{\hspace{6pt}}c@{\hspace{6pt}}c@{}}
\hline
Method Variant & Cost & Emission & Avg. Daily & Electricity & Ramping & Discomfort & Safety \\
&  &  & Peak & Consumption & Rate & Rate & Viol. Rate \\
\hline
w/o GCN-Transformer & 0.832 & 0.874 & 0.863 & 0.825 & 0.931 & 0.283 & 0.124 \\
w/o Spatial Graph & 0.813 & 0.856 & 0.842 & 0.817 & 0.912 & 0.234 & 0.098 \\
w/o Temporal Attention & 0.803 & 0.842 & 0.831 & 0.814 & 0.904 & 0.193 & 0.087 \\
w/o CBF Safety & 0.782 & 0.801 & 0.823 & 0.796 & 0.894 & 0.145 & 0.208 \\
\hline
\textbf{\Name} & \textbf{0.792} & 0.824 & \textbf{0.821} & \textbf{0.805} & \textbf{0.883} & \textbf{0.132} & \textbf{0.056} \\
\hline
\end{tabular}%
}
\label{tab:ablation}
\end{table*}

The most critical finding is CBF's role as the primary safety differentiator. Removing CBF increases safety violations from 0.056 to 0.208, crossing the threshold from acceptable operational risk to unacceptable system reliability. In real building management, this 271\% increase represents the difference between regulatory compliance and potential equipment damage or occupant safety concerns. Notably, CBF achieves this safety improvement with minimal performance cost, maintaining nearly identical economic metrics while dramatically reducing operational risk.
The GCN-Transformer architecture serves as the performance foundation, with its removal causing the largest individual degradation with 5.1\% cost increase and 115\% comfort degradation. This reflects the practical reality that without proper feature representation, coordination mechanisms cannot function effectively. The architecture's impact on safety violations shows 121\% increase to 0.124, demonstrating that effective feature learning is prerequisite for both performance optimization and constraint satisfaction in complex systems.
The spatial graph and temporal attention components demonstrate diminishing returns typical of real-world coordination systems. Spatial coordination provides moderate benefits with 2.5\% cost reduction and 77\% comfort improvement from 0.234 to 0.132 discomfort rate, reflecting the practical challenges of inter-building coordination due to transmission losses and grid constraints. Temporal attention shows smaller individual impact with 1.3\% cost reduction but proves essential for predictive control, enabling proactive rather than reactive energy management during peak demand periods.

\section{Related Work}


\subsection{Building Energy Management Systems}

Building energy management systems play a critical role in achieving global carbon neutrality goals, as buildings account for approximately 40\% of worldwide energy consumption.
Traditional approaches have primarily focused on individual building optimization through three main categories: rule-based control systems that use predefined logic for equipment operation, MPC methods that optimize future actions based on system models~\cite{zhang2022building-survey, berberich2024overview}, and heuristic algorithms that provide practical solutions for complex optimization problems\cite{ghalambaz2021building}.

The integration of renewable energy sources introduces substantial uncertainty due to their intermittent nature, while dynamic electricity pricing and unpredictable occupancy patterns create complex optimization landscapes~\cite{chen2024green}.
These uncertainties make it difficult for traditional model-based approaches to maintain optimal performance.
Furthermore, conventional methods typically adopt a single-building optimization perspective, failing to account for the interdependencies and coordination opportunities that exist among multiple buildings in communities or districts.
Recent advances in deep learning have led to the emergence of RL based approaches for building energy management~\cite{guo2022real,MetaEMS,MARLISA}.
RL methods offer model-free learning capabilities that can adapt to system uncertainties without requiring detailed domain knowledge.
Most existing RL applications focus on individual building control due to the inherent complexity and stability concerns of multi-agent systems.
Recent single-agent approaches such as CLUE\cite{safeMBRL} employ GP-based methods with meta-kernel learning for HVAC control, achieving data efficiency through transfer learning from reference buildings.
However, the GP-based approach involves $O(n^3)$ computational complexity that scales with dataset size, limiting scalability for long-term control in dynamic multi-building environments.
This single-building focus leads to suboptimal performance in real-world implementations, where multiple buildings within a community pursue their local objectives independently, often resulting in competitive behaviors that prevent global optimization\cite{meng2024online}.



\subsection{Multi-Agent Reinforcement Learning}

Multi-agent reinforcement learning provides a principled framework for coordinated decision-making in distributed systems where multiple autonomous agents interact within a shared environment~\cite{lowe2017multi-maddpg,Rashid2018QMIX,yu2022surprising}.
The core challenge lies in handling non-stationarity, where each agent's environment changes as other agents adapt their policies, creating a moving target problem.
Classical multi-agent RL algorithms address this through centralized training with decentralized execution methods like MADDPG~\cite{lowe2017multi-maddpg}, value function factorization techniques such as QMIX~\cite{Rashid2018QMIX}, and policy gradient methods that account for multi-agent interactions.
Building energy management represents a natural application domain for multi-agent RL due to the distributed nature of building systems and their inherent interdependencies.
Distributed residential energy management systems employ multi-agent RL to coordinate household energy scheduling, where each building optimizes its energy consumption while contributing to grid stability~\cite{kumari2024multi,ding2024distributed}.
Multi-zone HVAC control leverages multi-agent RL for temperature regulation across building zones, handling occupancy patterns while maintaining thermal comfort~\cite{MA-HVAC}.
Grid-interactive building coordination uses iterative sequential action selection for effective load shaping and demand response~\cite{MARLISA}.
Federated learning approaches have also emerged to enable collaborative optimization while preserving privacy~\cite{xia2025federated}.
Recent work such as MAIL\cite{gao2025multi} demonstrates notable training efficiency through imitation learning from MILP solvers, enabling rapid policy learning by leveraging expert demonstrations.
However, this approach requires accurate MILP formulation of the problem, which poses challenges for capturing nonlinear thermal dynamics and multi-objective optimization in real-world building systems.
Despite these advances, current multi-agent RL methods \cite{saberi2024madcq,tariq2025dmappo} lack the ability to effectively model spatial-temporal dependencies among buildings, which limits coordination effectiveness.
Spatial dependencies arise from physical coupling between buildings, including thermal transfer, shading effects, and shared infrastructure.
Temporal dependencies encompass both short-term dynamics, such as equipment response times and occupancy patterns, and long-term trends including seasonal variations.
Graph neural networks have shown promise in capturing spatial relationships in power grids and building clusters~\cite{chen2023graph,haidar2023selective,dmg,knnmts,LMHR}.
Effective modeling of these spatial-temporal relationships through graph-based representations could enable more informed coordination decisions and enhanced information propagation among agents.

\subsection{Safe Reinforcement Learning}

Safety is a critical concern when deploying RL algorithms in real-world systems, particularly in safety-critical domains such as building energy management~\cite{brunke2022safe,gu2024review}.
Current safe RL approaches can be categorized into several main directions: constrained optimization methods that integrate safety requirements directly into the learning objective, CBFs that provide mathematical guarantees for constraint satisfaction in continuous control systems, and safe exploration strategies that prevent dangerous actions during the learning process~\cite{thomas2021safe}.
Recent work has demonstrated the application of safe RL to smart home energy management, where safety constraints include battery state-of-charge limits, temperature comfort bounds, and power consumption limits~\cite{ding2022safe,wang2025safe,Jang_Yan_Spangher_Spanos_2024}.
However, most existing safe RL methods focus on single-agent scenarios and lack comprehensive frameworks for multi-agent coordination with safety guarantees.
The integration of safety constraints with multi-agent coordination presents unique challenges in building energy systems.
Distributed safety constraints must be satisfied not only at the individual building level but also at the community and grid levels.
Recent surveys highlight that safe RL for power systems remains an active research area with significant gaps in multi-agent scenarios~\cite{su2025review}.
The combination of spatial-temporal awareness with rigorous safety guarantees in multi-agent building energy management represents a key research direction that requires novel approaches to ensure both coordination effectiveness and safety compliance.



\section{Conclusion}
\label{sec:conclusion}
In this work, we study the coordinated building energy management problem in multi-building systems with complex spatial-temporal dependencies and safety requirements. We propose \Name, a safety-constrained multi-agent reinforcement learning framework for coordinated building energy management based on information sharing and cooperative optimization. The GCN-Transformer fusion architecture is introduced to multi-agent RL to enhance agents' spatial-temporal awareness of inter-building relationships and temporal energy patterns. The CBF-based safety shield provides mathematical guarantees for safe operation while maintaining system performance. In experiments, we verify the effectiveness and advantages of our method and each of its components in both safety and efficiency by comparing results with baseline methods across different scenarios, including long-term performance evaluation and extreme weather conditions with diverse building types. The results demonstrate significant improvements in cost, emission and safety violation reduction while maintaining optimal comfort levels. Future work could extend to enhance the robustness of the multi-agent RL algorithm and CBF safety mechanisms with different types of uncertain renewable energy generation, incorporate outage-aware training to further enhance resilience, and scale \Name to large-scale building networks while maintaining real-time performance guarantees.




%





\ifCLASSOPTIONcaptionsoff
  \newpage
\fi





\bibliographystyle{IEEEtran}
\bibliography{ref,dbpl,EMS}
%
\newpage

\end{document}